\documentclass[conference]{IEEEtran}

\IEEEoverridecommandlockouts                               
\usepackage[table]{xcolor}
\usepackage{graphicx} 
\usepackage{chngpage}
\usepackage{calc}
\usepackage{epsfig} 
\usepackage{mathptmx} 
\usepackage{amsmath} 
\usepackage{amssymb}  
\usepackage[hyphens]{url}
\usepackage{enumitem}
\usepackage{multirow}
\usepackage{algorithm}
\usepackage{algorithmic}
\usepackage{subcaption}
\usepackage{xcolor}
\usepackage{xspace}
\usepackage{nicefrac}
\usepackage{numprint}
\usepackage{printlen}
\usepackage{textcomp}
\usepackage{authblk}

\makeatletter
\def\@xfootnote[#1]{%
  \protected@xdef\@thefnmark{#1}%
  \@footnotemark\@footnotetext}
\makeatother

\DeclareMathAlphabet{\mathcal}{OMS}{cmsy}{m}{n}

\DeclareCaptionLabelFormat{custom}
{%
      #1 \textbf{(#2)}
}
\DeclareCaptionFormat{custom}
{%
    \small {#1#2 #3}
}

\begin{document}

\makeatletter
\def\endthebibliography{%
  \def\@noitemerr{\@latex@warning{Empty `thebibliography' environment}}%
  \endlist
}
\makeatother

\title{To ChatGPT, or not to ChatGPT: \\ That is the question!}
\author[*]{Alessandro Pegoraro}
\author[*]{Kavita Kumari}
\author[*]{Hossein Fereidooni}
\author[*]{Ahmad-Reza Sadeghi}
\affil[*]{System Security Lab, Technical University of Darmstadt, Germany}


\maketitle

\begin{abstract}
ChatGPT has become a global sensation.
As ChatGPT and other Large Language Models (LLMs) emerge, concerns of misusing them in various ways increase, such as disseminating fake news, plagiarism, manipulating public opinion, cheating, and fraud.
\noindent Hence, distinguishing AI-generated from human-generated becomes increasingly essential. Researchers have proposed various detection methodologies, ranging from basic binary classifiers to more complex deep-learning models. Some detection techniques rely on statistical characteristics or syntactic patterns, while others incorporate semantic or contextual information to improve accuracy.
\\
\noindent The primary objective of this study is to provide a comprehensive and contemporary assessment of the most recent techniques in ChatGPT detection.  Additionally, we evaluated other AI-generated text detection tools that do not specifically claim to detect ChatGPT-generated content to assess their performance in detecting ChatGPT-generated content. 
For our evaluation we have curated a benchmark dataset consisting of prompts from ChatGPT and humans, including diverse questions from medical, open Q\&A, and finance domains and user-generated responses from popular social networking platforms. The dataset serves as a reference to assess the performance of various techniques in detecting ChatGPT-generated content. Our evaluation results demonstrate that none of the existing methods can effectively detect ChatGPT-generated content.

\end{abstract}

\bigskip
\section{Introduction}
\label{sec:Intro}

\noindent ChatGPT developed by OpenAI has garnered significant attention and sparked extensive discourse in the Natural Language Processing (NLP) community and several other fields. ChatGPT is an AI chatbot introduced by OpenAI in November 2022. It utilizes the power of OpenAI's LLMs belonging to the \textit{GPT-3.5} and \mbox{\textit{GPT-4}} families. However, ChatGPT is not a simple extension of these models. Instead, it has undergone a fine-tuning process utilizing supervised and reinforcement learning techniques based on Human Feedback \cite{christiano2017deep, lambert2022illustrating}. This approach to transfer learning has allowed ChatGPT to learn from existing data and optimize its performance for conversational applications. It has also facilitated ChatGPT's exceptional performance in various challenging NLP tasks~\cite{biswas2023chatgpt, dowling2023chatgpt, omar2023chatgpt, susnjak2023applying, yeo2023assessing}.
The media's promotion of ChatGPT has resulted in a chain of reactions, with news and media companies utilizing it for optimal content creation, teachers and academia using it to prepare course objectives and goals, and individuals using it to translate content between languages. Unfortunately, as is often the case with such technologies, misuse has also ensued. Students are employing it to generate their projects and coding assignments \cite{bleumink2023keeping, cotton2023chatting}, while scholars are utilizing it to produce papers \cite{gao2022comparing}. Malicious actors use it to propagate fake news on social media platforms \cite{hacker2023regulating, de2023chatgpt}, and educational institutions are employing it to provide mental health education to students without their consent \cite{mentalhealth}, among other uses. Furthermore, ChatGPT has the potential to generate seemingly realistic stories that could deceive unsuspecting readers \cite{bang2023multitask, shen2023chatgpt}. Hence, developing an efficient detection algorithm capable of distinguishing AI-generated text, particularly ChatGPT, from human-generated text has attracted many researchers. 

\noindent In general, detecting AI-generated text using machine learning concerns two types: black-box and white-box detection. Black-box detection relies on API-level access to language models, limiting its capability to detect synthetic texts~\cite{tang2023science}. This type involves data collection, feature extraction, and building a classifier for detection. It includes simple classifiers such as binary logistic regression~\cite{solaiman2019release}. In contrast, white-box detection has full access to language models, enabling control of the model's behavior and traceable results~\cite{tang2023science}. It includes zero-shot detection~\cite{mitchell2023detectgpt, solaiman2019release, zellers2019defending} that leverages pre-trained generative models like \textit{GPT-2}~\cite{solaiman2019release} or \textit{Grover}~\cite{zellers2019defending} and pre-trained language models fine-tuned for the task. 

\noindent A large body of research is attributed to building detectors for the text generated by AI bots~\cite{cotton2023chatting, gao2022comparing, gehrmann2019gltr, perplexityAnalysis, khalil2023will, kumarage2023stylometric, kushnareva2021artificial, mitchell2023detectgpt, GPTZero, solaiman2019release, zellers2019defending, AITextDetector}. Furthermore, some claim that their AI-text detector can distinguish the ChatGPT-generated text from the human-generated text~\cite{originality, aicontentdetector, bleumink2023keeping, copyleaks, DAG, Huggingface, Frohling2021feature, guo2023close, mitrovic2023chatgpt, AITextClassifier, writefull, WriterAIContentDetector}. On that account, our motivation is to test all the tools (generalized AI-text detectors plus detectors targeting ChatGPT-generated text) against a benchmark dataset (Section \ref{sub:dataset}), comprising of ChatGPT prompts and human responses, spanning different domains. We will elaborate on each tool and its functionality in the following section. 

\smallskip
\noindent Our goals in this paper are as follows:

\begin{itemize}
\item We explore the research conducted on AI-generated text detection focusing on ChatGPT detection. We outline different white-box and black-box detection schemes proposed in the literature. We also explore detection schemes in education, academic/scientific writing, and the detection tools available online. 

\item We evaluate the effectiveness of various tools that aim to distinguish ChatGPT-generated from human-generated responses and compare their accuracy and reliability. Our assessment includes tools that claim to detect ChatGPT prompts and other AI-generated text detection tools that do not target ChatGPT-generated content. This evaluation's primary objective is to gauge these tools' effectiveness in detecting ChatGPT-generated content. 
Our analysis reveals that the most effective online tool for detecting generated text can only achieve a success rate of less than 50\%, as depicted in Table~\ref{tab:summary}.

\item Our research aims to inspire further inquiry into this critical area of study and promote the development of more effective and accurate detection methods for AI-generated text. Further, our findings underscore the importance of thorough testing and verification when assessing AI detection tools.
\end{itemize}

\bigskip
\section{Related works}
\label{sec:analysis}
\noindent This section provides an overview of current research on distinguishing AI-generated text from human-generated text. To categorize most automated machine learning-based detection methods for synthetic texts, we followed OpenAI's classification \cite{solaiman2019release}, which divides these methods into three main categories, which are: i) Simple Classifiers \cite{guo2023close, solaiman2019release}, ii) Zero-shot detection techniques \cite{kushnareva2021artificial, mitchell2023detectgpt, zellers2019defending}, and iii) Fine-tuning based detection \cite{mitrovic2023chatgpt}. Simple Classifiers fall under the category of black-box detection techniques, whereas zero-shot and fine-tuning-based detection techniques come under the umbrella of white-box detection techniques. There exists other approaches that do not fit into these three categories; however, they are still significant and merit consideration. These alternative methods include testing ChatGPT-generated text against various plagiarism tools \cite{khalil2023will}, designing a Deep Neural Network-based AI detection tool \cite{bleumink2023keeping}, a sampling-based approach \cite{gehrmann2019gltr}, and online detection tools~\cite{originality, aicontentdetector, copyleaks, Huggingface, perplexityAnalysis, AITextClassifier, GPTZero, writefull, AITextDetector}. In the following sections, we will analyze the existing approaches belonging to the aforementioned categories and alternative methods, focusing on their effectiveness in detecting AI-generated text.

\smallskip
\subsection{Simple Classifiers}
\label{sub:simple}
\noindent OpenAI \cite{solaiman2019release}, an artificial intelligence research company, analyzed the human detection and automated ML-based detection of synthetic texts. For the human detection of the synthetic datasets, authors showed that the models trained on the \textit{GPT-2} datasets tend to increase the perceived humanness of \textit{GPT-2} generated text. Hence, they tested a simple logistic regression model, zero-shot detection model (explained in Section \ref{sub:zero}), and fine-tune-based detection model (described Section in \ref{sub:tuning}). The simple logistic regression model was trained on TF-IDF (Term Frequency-Inverse Document Frequency), unigram, and bigram features and later analyzed at different generation strategies and model parameters. It was found that the simple classifiers can work correctly up to an accuracy of 97\%. However, detecting shorter outputs is more complicated than detecting more extended outputs for these models.
\noindent Guo et al. \cite{guo2023close} conducted human evaluations and compared datasets generated by ChatGPT and human experts, analyzing linguistic and stylistic characteristics of their responses and highlighting differences between them. The authors then attempted to detect whether the text was ChatGPT-generated or human-generated by deploying: first, a simple logistic regression model on \mbox{\textit{GLTR} Test-2} features and, second, a pre-trained deep classifier model based on \textit{RoBERTa}~\cite{liu2019roberta} for single-text and Q\&A detection (as explained in Section \ref{sub:tuning}). However, the proposed detection models are ineffective in detecting ChatGPT-generated text from human-generated text due to the highly unbalanced corpus of datasets, which did not capture all the text-generating styles of ChatGPT.
\noindent Another study by Kushnareva et al. \cite{kushnareva2021artificial} utilized \textit{Topological Data Analysis} (TDA) to extract three types of interpretable topological features, such as the number of connected components, the number of edges, and the number of cycles present in the graph, for artificial text recognition. The authors then trained a logistic regression classifier with these features and tested the approach on datasets from 
WebText \& \textit{GPT-2}, Amazon Reviews \& \textit{GPT-2}, and RealNews \& \textit{GROVER}~\cite{zellers2019defending}. However, this approach is unlikely to be effective for ChatGPT, as it was not tested on that specific model.

\subsection{Zero-shot detection techniques}
\label{sub:zero}
\noindent OpenAI \cite{solaiman2019release} has also developed a \textit{GPT-2} detector using a 1.5 billion parameter \textit{GPT-2} model that can identify the top 40 generated outputs with an accuracy of 83\% to 85\%. However, when the model was fine-tuned to the Amazon reviews dataset, the accuracy dropped to 76\%. In a different study \cite{mitchell2023detectgpt}, the authors explored the Zero-shot detection of AI-generated text and deployed an online detection tool (\textit{DetectGPT}) to distinguish \textit{GPT-2} generated text from the human-generated text. They used the generative model's log probabilities to achieve this. The authors experimented and demonstrated that AI-generated text occupies the negative curvature regions of the model's log probability function. However, it should be noted that the authors assumed that one could evaluate the log probabilities of the model(s) under consideration, which may not always be possible. Moreover, as mentioned by the authors, this approach is only practical for \textit{GPT-2} prompts. 

\noindent Zellers et al. \cite{zellers2019defending} utilized a transformer identical to the one used for \textit{GPT-2}, except that they used nucleus sampling instead of top-k sampling to select the next word during text generation. The model they developed, known as \textit{Grover}, can generate text such as fake news and detect its own generated text. It is also available online. Authors used \textit{Grover}, \textit{GPT-2} (124M or 355M ), BERT (BERT-Base or BERT-Large), and FastText verification tools to classify the news articles generated by \textit{Grover}. They proved that \textit{Grover} is the best among the previously mentioned detector to verify its self-generated fake news. However, it's not visible that it will work for the text generated by the \textit{GPT} models. Also, it has been shown that the bi-directional transformer model \textit{RoBERTa} outperforms \textit{Grover} models with equivalent parameter size in detecting \textit{GPT-2} texts \cite{solaiman2019release}.

\subsection{Fine-tuning based detection}
\label{sub:tuning}
\noindent In \cite{solaiman2019release}, the authors conducted experiments to fine-tune pre-trained language models for detecting AI-generated texts by basing the classifiers on \textit{$RoBERTa_{BASE}$} and \textit{$RoBERTa_{LARGE}$}. They found that fine-tuning \textit{RoBERTa} consistently outperformed fine-tuning an equivalent capacity \mbox{\textit{GPT-2}} model. However, the approach could not detect text generated by ChatGPT, as demonstrated in \cite{rudolph2023chatgpt}. In a separate study, Mitrovic et al. \cite{mitrovic2023chatgpt} investigated the feasibility of training an ML model to distinguish between queries generated by \mbox{ChatGPT} and those generated by humans. The authors attempted to detect ChatGPT-generated two-line restaurant reviews using a framework based on \textit{DistilBERT}, a lightweight model trained using \textit{BERT}, which was fine-tuned using a Transformer-based model. Additionally, the predictions made by the model were explained using the \textit{SHAP} method. The authors concluded that an ML model could not successfully identify texts generated by ChatGPT. Guo at al. \cite{guo2023close} also developed a pre-trained deep classifier model based on \textit{RoBERTa} for single-text and Q\&A detection. The limitations are the same as described in Section \ref{sub:simple}.

\begin{table*}[t]
\renewcommand{\arraystretch}{1.1}
\begin{minipage}{\textwidth}
    \caption{Summary of analyzed papers}
    \label{tab:summary}
    \centering
        \begin{tabular}{|l|c|cccc|c|c|c|c|}
\hline
\multicolumn{1}{|c|}{\multirow{2}{*}{Approach}} & \multicolumn{1}{c|}{\multirow{2}{*}{Published in}} & \multicolumn{4}{c|}{Target Model}                                                               & \multirow{2}{*}{\begin{tabular}[c]{@{}c@{}}Publicly\\ Available\end{tabular}} & \multirow{2}{*}{Free/Paid} & \multirow{2}{*}{\begin{tabular}[c]{@{}c@{}}ChatGPT detc.\\  Capability (\textit{TPR}\%)\end{tabular}} & \multirow{2}{*}{\begin{tabular}[c]{@{}c@{}}Human-text detc. \\ Capability (\textit{TNR}\%)\end{tabular}} \\ 
\multicolumn{1}{|c|}{} & \multicolumn{1}{c|}{}                          & \multicolumn{1}{c}{Grover} & \multicolumn{1}{c}{GPT-2} & \multicolumn{1}{c}{GPT-3} & \multicolumn{1}{c|}{ChatGPT\footnote[*]{GPT 3.5 and above.}} &                                                                               &                            &                                                                                          &                                                                                             \\ \hline
Kumarage et al.~\cite{kumarage2023stylometric}  & \multicolumn{1}{c|}{2023}                                & \multicolumn{1}{c|}{}       & \multicolumn{1}{c|}{\checkmark}   & \multicolumn{1}{c|}{}      &         & \checkmark                                                                           & Free             & 23.3                                                                                     & 94.7                                                                                        \\ \hline
Bleumink et al.~\cite{bleumink2023keeping}  & \multicolumn{1}{c|}{2023}                               & \multicolumn{1}{c|}{}       & \multicolumn{1}{c|}{}      & \multicolumn{1}{c|}{\checkmark}   & \checkmark     & \checkmark                                                                           & Paid                   & 13.4                                                                                     & 95.4                                                                                        \\ \hline
ZeroGPT~\cite{AITextDetector}          & \multicolumn{1}{c|}{2023}                               & \multicolumn{1}{c|}{}       & \multicolumn{1}{c|}{}      & \multicolumn{1}{c|}{}      & \checkmark     & \checkmark                                                                           & Paid                   & 45.7                                                                                     & 92.2                                                                                        \\ \hline
OpenAI  Classifier~\cite{AITextClassifier}  & \multicolumn{1}{c|}{2023}                            & \multicolumn{1}{c|}{}       & \multicolumn{1}{c|}{}      & \multicolumn{1}{c|}{}      & \checkmark     & \checkmark                                                                           & Free                  & 31.9                                                                                     & 91.8                                                                                        \\ \hline
Mitchell et al.~\cite{mitchell2023detectgpt}    & \multicolumn{1}{c|}{2023}                             & \multicolumn{1}{c|}{}       & \multicolumn{1}{c|}{\checkmark}   & \multicolumn{1}{c|}{}      &         & \checkmark                                                                           & Free                  & 18.1                                                                                     & 80.0                                                                                       \\ \hline
GPTZero~\cite{GPTZero}           & \multicolumn{1}{c|}{2023}                              & \multicolumn{1}{c|}{}       & \multicolumn{1}{c|}{\checkmark}   & \multicolumn{1}{c|}{\checkmark}   & \checkmark     & \checkmark                                                                           & Paid                   & 27.3
& 93.5
\\ \hline
Hugging Face~\cite{Huggingface}      & \multicolumn{1}{c|}{2023}                              & \multicolumn{1}{c|}{}       & \multicolumn{1}{c|}{}      & \multicolumn{1}{c|}{}      & \checkmark     & \checkmark                                                                           & Free                   & 10.7                                                                                      & 62.9                                                                                        \\ \hline
Guo et al.~\cite{guo2023close}  & \multicolumn{1}{c|}{2023}                                    & \multicolumn{1}{c|}{}       & \multicolumn{1}{c|}{}      & \multicolumn{1}{c|}{}      & \checkmark     & \checkmark                                                                           & Free                   & 47.3
& 98.0
\\ \hline
Perplexity (PPL)~\cite{perplexityAnalysis} & \multicolumn{1}{c|}{2023}                                    & \multicolumn{1}{c|}{}       & \multicolumn{1}{c|}{}      & \multicolumn{1}{c|}{}      & \checkmark     & \checkmark                                                                           & Free                   & 44.4
& 98.3
\\ \hline
Writefull GPT~\cite{writefull}  & \multicolumn{1}{c|}{2023}                                 & \multicolumn{1}{c|}{}       & \multicolumn{1}{c|}{}      & \multicolumn{1}{c|}{\checkmark}   & \checkmark     & \checkmark                                                                           & Paid                   & 21.6                                                                                      & 99.3                                                                                        \\ \hline
Copyleaks~\cite{copyleaks}       & \multicolumn{1}{c|}{2023}                                & \multicolumn{1}{c|}{}       & \multicolumn{1}{c|}{}      & \multicolumn{1}{c|}{\checkmark}   & \checkmark     & \checkmark                                                                           & Paid                   & 22.9                                                                                      & 92.1                                                                                        \\ \hline
Cotton et al.~\cite{cotton2023chatting}     & \multicolumn{1}{c|}{2023}                              & \multicolumn{1}{c|}{}       & \multicolumn{1}{c|}{}      & \multicolumn{1}{c|}{\checkmark}   & \checkmark     & $\times$                                                                         &           -                 & -                                                                                         & -                                                                                            \\ \hline
Khalil et al.~\cite{khalil2023will}     & \multicolumn{1}{c|}{2023}                              & \multicolumn{1}{c|}{}       & \multicolumn{1}{c|}{}      & \multicolumn{1}{c|}{}      & \checkmark     & $\times$                                                                         &            -                & -                                                                                         & -                                                                                            \\ \hline
Mitrovic et al.~\cite{mitrovic2023chatgpt}    & \multicolumn{1}{c|}{2023}                             & \multicolumn{1}{c|}{}       & \multicolumn{1}{c|}{\checkmark}   & \multicolumn{1}{c|}{}      & \checkmark     & $\times$                                                                           &               -             & -                                                                                         & -                                                                                            \\ \hline

Content at Scale~\cite{aicontentdetector}  & \multicolumn{1}{c|}{2022}                              & \multicolumn{1}{c|}{}       & \multicolumn{1}{c|}{\checkmark}   & \multicolumn{1}{c|}{\checkmark}   & \checkmark     & \checkmark                                                                           & Paid                   
& 38.4
& 79.8
\\ \hline

Orignality.ai~\cite{originality}  & \multicolumn{1}{c|}{2022}                              & \multicolumn{1}{c|}{}       & \multicolumn{1}{c|}{}   & \multicolumn{1}{c|}{\checkmark}   & \checkmark     & $\times$                                                                          & Paid                   & 7.6                                                                                     & 95.0                                                                                        \\ \hline

Writer AI Detector~\cite{WriterAIContentDetector}   & \multicolumn{1}{c|}{2022}                           & \multicolumn{1}{c|}{}       & \multicolumn{1}{c|}{}      & \multicolumn{1}{c|}{\checkmark}   & \checkmark     & \checkmark                                                                           & Paid                   
& 6.9
& 94.5
\\ \hline

Draft and Goal~\cite{DAG}        & \multicolumn{1}{c|}{2022}                          & \multicolumn{1}{c|}{}       & \multicolumn{1}{c|}{}      & \multicolumn{1}{c|}{\checkmark}   & \checkmark     & \checkmark                                                                           & Free                  & 23.7                                                                                      & 91.1                                                                                        \\ \hline

Gao et al.~\cite{gao2022comparing}    & \multicolumn{1}{c|}{2022}                                  & \multicolumn{1}{c|}{}       & \multicolumn{1}{c|}{}      & \multicolumn{1}{c|}{}      & \checkmark     & $\times$                                                                           &           -                 & -                                                                                         & -                                                                                            \\ \hline

Fröhling et al.~\cite{Frohling2021feature}   & \multicolumn{1}{c|}{2021}                              & \multicolumn{1}{c|}{\checkmark}    & \multicolumn{1}{c|}{\checkmark}   & \multicolumn{1}{c|}{\checkmark}   &         & \checkmark                                                                           & Free             & 27.8                                                                                     & 89.2                                                                                        \\ \hline

Kushnareva et al.~\cite{kushnareva2021artificial}   & \multicolumn{1}{c|}{2021}                            & \multicolumn{1}{c|}{\checkmark}    & \multicolumn{1}{c|}{\checkmark}   & \multicolumn{1}{c|}{}      &         & \checkmark                                                                           & Free             & 25.1                                                                                     & 96.3                                                                                        \\ \hline

Solaiman et al.~\cite{solaiman2019release}     & \multicolumn{1}{c|}{2019}                            & \multicolumn{1}{c|}{}       & \multicolumn{1}{c|}{\checkmark}   & \multicolumn{1}{c|}{}      &         & \checkmark                                                                           & Free                  & 7.2                                                                                      & 96.4                                                                                        \\ \hline

Gehrmann et al.~\cite{gehrmann2019gltr}   & \multicolumn{1}{c|}{2019}                              & \multicolumn{1}{c|}{}       & \multicolumn{1}{c|}{\checkmark}   & \multicolumn{1}{c|}{}      &         & \checkmark                                                                           & Free                  
& 32.0   
& 98.4        
\\ \hline

Zellers et al.~\cite{zellers2019defending}  & \multicolumn{1}{c|}{2019}                                & \multicolumn{1}{c|}{\checkmark}    & \multicolumn{1}{c|}{}      & \multicolumn{1}{c|}{}      &         & \checkmark                                                                           & Free                  & 43.1                                                                                     & 91.3                                                                                        \\ \hline
\end{tabular}
\end{minipage}
\end{table*}

\subsection{Other approaches proposed in literature}
\label{sub:other}
\noindent Extensive academic research has investigated the adverse impact of ChatGPT on education, which has shown that students and scholars can use ChatGPT to engage in plagiarism. To address this issue, several existing tools, such as \textit{RoBERTa}, \textit{Grover}, or \textit{GPT-2}, have been utilized to check the uniqueness of educational content against ChatGPT-generated text. In \cite{bleumink2023keeping}, the authors proposed a transformer-based model named \textit{AICheatCheck}, a web-based AI detection tool designed to identify whether a human or ChatGPT generated a given text. \textit{AICheatCheck} examines a sentence or group of sentences for patterns to determine their origin. 
The authors used the data collected by Guo et al. \cite{guo2023close} (with its limitations) and from the education field. Also, it is not specified in the paper on what basis or on what features \textit{AICheatCheck} can achieve high accuracy.  

\noindent The study in \cite{khalil2023will}, evaluates the effectiveness of two popular plagiarism-detection tools, \textit{iThenticate} and \textit{Turnitin}, in detecting plagiarism concerning 50 essays generated by ChatGPT. The authors also compared ChatGPT's performance against itself and found it more effective than traditional plagiarism detection tools. Another study by Gao et al. \cite{gao2022comparing} aimed to compare ChatGPT-generated academic paper abstracts using a \textit{GPT-2} Output Detector \textit{RoBERTa}, a plagiarism checker, and human review. The authors collected ten research abstracts from five high-impact medical journals and then used ChatGPT to output research abstracts based on their titles and journals. However, the tool used in the study is not available online for verification. In recent work, Cotton et al. \cite{cotton2023chatting} investigate the pros and cons of using ChatGPT in the academic field, particularly concerning plagiarism. In a different work, authors in \cite{gehrmann2019gltr} utilized the distributional properties of the underlying text used for the model. They deployed a tool called \textit{GLTR} that highlights the input text in different colors to determine its authenticity. \textit{GLTR} was tested on the prompts from \textit{GPT-2} 1.5B parameter model \cite{radford2019language} and human-generated articles present on social media platforms. The authors also conducted a human study, asking the students to identify fake news from real news.

\subsection{Online tools}
\label{sub:oonlinetool}
\noindent Below, we examine multiple online tools and delineate their brief functionality.
    \begin{enumerate}[leftmargin=*]
        \item \textit{Stylometric Detection of AI-generated Text}~\cite{kumarage2023stylometric}: This tool utilizes stylometric signals to examine the writing style of a text by identifying patterns and features that are unique to them. These signals are then extracted from the input text to enable sequence-based detection of AI-generated tweets.
        \item \textit{ZeroGPT}~\cite{AITextDetector}: This tool is specifically developed to detect OpenAI text but has limited capabilities with shorter text.
        \item \textit{OpenAI Text Classifier}~\cite{AITextClassifier}: This fine-tuned GPT model developed by OpenAI predicts the likelihood of a text being AI-generated from various sources, including \textit{ChatGPT}. However, the tool only works with 1000 characters or more and is less reliable in determining if a text was artificially generated.
        \item \textit{GPTZero}~\cite{GPTZero}: This classification model determines whether a document was written by an LLM, providing predictions at a sentence, paragraph, and document level. However, it mainly works for content in the English language and only allows text with a character count between 250 and 5000 characters.
        \item \textit{Hugging Face}~\cite{Huggingface}: This tool was released by Hugging Face for detecting text generated by \textit{ChatGPT}. However, it tends to over-classify text as being \textit{ChatGPT}-written.
        \item \textit{Perplexity (PPL})~\cite{perplexityAnalysis}: The perplexity (PPL) is a widely employed metric for assessing the efficacy of Large language models (LLM). It is calculated as the exponential of the negative average log-likelihood of text under the LLM. A lower PPL value implies that the language model is more confident in its predictions. LLMs are trained on vast text corpora, enabling them to learn common language patterns and text structures. Consequently, PPL can be utilized to gauge how effectively a given text conforms to such typical characteristics.
        \item \textit{Writefull GPT Detector}~\cite{writefull}: It is primarily used for detecting plagiarism, this tool can identify if a piece of text is generated by \textit{GPT-3} or \textit{ChatGPT}. However, the tool's percentage-based system for determining whether the text was created by AI has a degree of uncertainty for both samples generated by humans and those generated by ChatGPT.
        \item \textit{Copyleaks}~\cite{copyleaks}: This tool claims to detect if a text is generated by \textit{GPT-3}, \textit{ChatGPT}, humans, or a combination of humans and AI. The tool accepts only text with 150 or more characters.
        \item \textit{Content at Scale}~\cite{aicontentdetector}:
        This is an online tool available to detect text generated by \textit{ChatGPT}. However, the tool can only analyze samples with 25000 characters or less.
        \item \textit{Originality.ai}~\cite{originality}: This paid tool is designed to work with \textit{GPT-3}, \textit{GPT 3.5} (DaVinci-003), and \textit{ChatGPT} models. However, the tool only works with 100 words or more and is prone to classify \textit{ChatGPT}-generated content as real.
        \item \textit{Writer AI Content Detector}~\cite{WriterAIContentDetector}: This tool is designed to work with \textit{GPT-3} and \textit{ChatGPT} models. However, its limitation restricts the amount of text that can be checked on each experiment to a maximum of 1500 characters.
        \item \textit{Draft and Goal}~\cite{DAG}: This tool is intended for detecting content generated by \textit{GPT-3} and \textit{ChatGPT} models, and it is equipped to perform detection in both English and French. However, it has a requirement that the input text should be at least 600 characters or longer to work effectively.
    \end{enumerate}

\section{Evaluation}
\label{sec:test}
\noindent This section evaluates publicly available literature, tools, or codes that can differentiate between AI-generated and human-generated responses. \textit{Our primary focus is on the tools claiming to detect ChatGPT-generated content}. However, we also evaluate (to the best of our abilities) the performance of other AI-generated text detection tools that do not make explicit claims about detecting ChatGPT-generated content on ChatGPT prompts. To assess the effectiveness of these tools, we employ a benchmark dataset (Section \ref{sub:dataset}) that comprises prompts from ChatGPT and humans. Then, we measure the detection capabilities of these tools on both ChatGPT-generated and human-generated content and present the results in Table~\ref{tab:summary}.

\subsection{Benchmark Dataset}
\label{sub:dataset}
\noindent We utilized the inquiry prompts proposed by Guo et al.~\cite{guo2023close} through the OpenAI API\footnote{\url{https://openai.com/blog/introducing-chatgpt-and-whisper-apis}} to generate a benchmark dataset. This dataset comprises 58,546 responses generated by humans and 72,966 responses generated by the ChatGPT model, resulting in 131,512 unique samples that address 24,322 distinct questions from various fields, including medicine, open-domain, and finance. Furthermore, the dataset incorporates responses from popular social networking platforms, which provide a wide range of user-generated perspectives. To assess the similarity between human-generated and ChatGPT-generated responses, we employed the sentence transformer \mbox{\textit{all-MiniLM-L6-v2}}\footnote{\url{https://huggingface.co/sentence-transformers/all-MiniLM-L6-v2}}. Then, we selected responses with the highest and lowest levels of similarity to assemble a benchmark dataset, which was reduced to approximately 10\% of the primary dataset that we generated. This benchmark dataset serves as a standardized reference for evaluating the ability of different techniques to detect ChatGPT-generated content.

\subsection{Evaluation Metrics} 
\noindent To measure and compare the effectiveness of each approach, we utilized the following metrics:
\begin{itemize}
\item \textit{True Positive Rate (TPR): } This metric represents the tool's sensitivity in detecting text that ChatGPT generates. True Positive ($TP$) is the total number of correctly identified samples, while we consider False Negative ($FN$) the number of samples not classified as generated text or incorrectly identified as human text. Therefore, $TPR = \frac{TP}{TP+FN}$.
\item \textit{True Negative Rate (TNR):} This metric indicates the tool's specificity in detecting human-generated texts. True Negatives ($TN$) is the total number of correctly identified samples, while False Positives ($FP$) is the number of samples incorrectly classified as being produced by ChatGPT. Therefore, $TNR = \frac{TN}{TN+FP}$.
\end{itemize}

\subsection{Evaluated Tools and Algorithms} 
  We evaluate several tools and algorithms summarized in Section \ref{sec:analysis}. Table \ref{tab:summary} outlines the detection capability of these tools for the ChatGPT-generated content in terms of \textit{TPR} and \textit{TNR}. We can observe that none of the evaluated approaches can consistently detect the ChatGPT-generated text. Analysis reveals that the most effective online tool for detecting generated text can only achieve a success rate of less than 50\%, as depicted in Table \ref{tab:summary}.




\section{Conclusion}
\label{sec:conc}

\noindent 
This study delved into the various methods employed for detecting ChatGPT-generated text. Through a comprehensive review of the literature and an examination of existing approaches, we assess the ability of these techniques to differentiate between responses generated by ChatGPT and those produced by humans. Furthermore, our study includes testing and validating online detection tools and algorithms utilizing a benchmark dataset that covers various topics, such as finance and medicine, and user-generated responses from popular social networking platforms. 
Our experiments highlight ChatGPT's exceptional ability to deceive detectors and further indicate that most of the analyzed detectors are prone to classifying any text as human-written, with a general high $TNR$ of 90\% and low $TPR$. These findings have significant implications for enhancing the quality and credibility of online discussions.
Ultimately, our results underscore the need for continued efforts to improve the accuracy and robustness of text detection techniques in the face of increasingly sophisticated AI-generated content.

\bibliographystyle{plain}
\bibliography{reference}

\end{document}